\documentclass[twocolumn,letterpaper,10pt]{article}
\usepackage{kuz-en}
\usepackage{graphicx}				% include graphics
\usepackage[utf8]{inputenc}  				% for czech diacritics (used mainly in names)
\usepackage[round,authoryear]{natbib}
\usepackage[english]{babel}					% change english titles to czech
\usepackage[labelfont=bf]{caption}			% figure captions in bold
\usepackage[colorlinks=false]{hyperref} 	% hyperref support

\usepackage{enumitem}

\usepackage{lipsum}

\makeatletter
\def\convertto#1#2{\strip@pt\dimexpr #2*65536/\number\dimexpr 1#1}
\makeatother

\begin{document}

\date{}
\title{The encoding of proprioceptive inputs in the brain: knowns and unknowns from a robotic perspective}

\author{
\vspace{1ex}
\textbf{Matěj Hoffmann} \\ % another author from the same isntitution goes here
Istituto Italiano di Tecnologia\\
Via Morego 30 \\
16163 Genova, Italy \\
Email: matej.hoffmann@iit.it \\
\and
\vspace{1ex}
\textbf{Naďa Bednárová} \\ % another author from the same isntitution goes here
Fakulta Elektrotechnická, ČVUT v Praze \\
Zikova 1903/4\\
166 36 Praha \\
Email: nada.bednarova@gmail.com \\
}
% \and
% \vspace{1ex}
% \textbf{ďalší autor (z inej inštitúcie)}\\
% Ďalšia prípadná adresa\\
% Adresa inej inštitúcie\\
% Email: dalsi@fmph.uniba.sk \\
%}

\maketitle 

\thispagestyle{empty}

%%%%%%%%%%%%%%%%%%%%%%%%%%%%%%%%%%%%%%%%%%%%%%%%%%%%%%%%%%%%%%%%%%%%%%%% ABSTRACT
\noindent
{ \begin{center} \bf\normalsize Abstract\end{center}}
{
\noindent
Somatosensory inputs can be grossly divided into tactile (or cutaneous) and proprioceptive -- the former conveying information about skin stimulation, the latter about limb position and movement. The principal proprioceptors are constituted by muscle spindles, which deliver information about muscle length and speed. In primates, this information is relayed to the primary somatosensory cortex and eventually the posterior parietal cortex, where integrated information about body posture (postural schema) is presumably available. However, coming from robotics and seeking a biologically motivated model that could be used in a humanoid robot, we faced a number of difficulties. First, it is not clear what neurons in the ascending pathway and primary somatosensory cortex code. To an engineer, joint angles would seem the most useful variables. However, the lengths of individual muscles have nonlinear relationships with the angles at joints. Kim et al. (Neuron, 2015) found different types of proprioceptive neurons in the primary somatosensory cortex -- sensitive to movement of single or multiple joints or to static postures. Second, there are indications that the somatotopic arrangement ("the homunculus") of these brain areas is to a significant extent learned. However, the mechanisms behind this developmental process are unclear. We will report first results from modeling of this process using data obtained from body babbling in the iCub humanoid robot and feeding them into a Self-Organizing Map (SOM). Our results reveal that the SOM algorithm is only suited to develop receptive fields of the posture-selective type. Furthermore, the SOM algorithm has intrinsic difficulties when combined with population code on its input and in particular with nonlinear tuning curves (sigmoids or Gaussians).  
}

%%%%%%%%%%%%%%%%%%%%%%%%%%%%%%%%%%%%%%%%%%%%%%%%%%%%%%%%%%%%%%%%%%%%%%%% INTRO
\section{Introduction}
Although proprioception does not belong to the commonly held "five senses" (sight, hearing, smell, touch, taste) view that dates back to Aristotle, it is, along with sense of balance for example, of outmost importance for our everyday existence. What is proprioception actually? The term has its origin in the work of \citet{Sherrington1906proprio} who stated: ``In muscular receptivity we see the body itself acting as a stimulus to its own receptors---the proprioceptors". In short, we will equate it with \textit{sense of limb position and movement} (see \citep{Proske2012} for an extensive survey). The reason why proprioceptive sensations are receiving so little attention in our everyray experience is mainly because we are largely unaware of them, which in turn can be attributed to their predictability. Unlike ``exteroceptors", such as sight or hearing that focus on events in the external environment that are not fully under our control, the position and movement of our limbs is something that is typically under our control and thus, these sensations typically receive less (conscious) attention \citep{Proske2012}. A related term is \textit{kinesthesia}, which strictly means movement sense, but has been sometimes equated with proprioception.

%%%%%%%%%%%%%%%%%%%%%%%%%%%%%%%%%%%%%%%%%%%%%%%%%%%%%%%%%%%%%%%%%%%%%%%% CHAPTER 2
\section{Representation of proprioception in biology}

\subsection{Receptors}

The key receptors in proprioception are muscle and skeletal mechanoreceptors: muscle spindles, Golgi tendon organs, and joint capsule mechanoreceptors---an overview is provided in Table \ref{tab:proprio_receptors}. They sense different physical quantities: while the muscle spindles and joint capsule mechanoreceptors are sensing kinematic variables (position and velocity), the Golgi tendons are sensing dynamic (or kinetic) quantities (forces). In addition, stretch-sensitive receptors in the skin (Ruffini endings, Merkel cells in hairy skin, and field receptors) also signal postural information \citet[p. 443]{Kandel2000}.\footnote{Muscle stretch can roughly be equated with muscle length.} However, in summary, ``the modern view has muscle spindles as the principal proprioceptors" \citep{Proske2012}.

\begin{table*}[t]
	\centering
	\begin{tabular}{|l||c|c|}
	\hline
	Receptor & Fiber group / name & Submodality \\ \hline \hline
	Muscle spindle primary & $A\alpha / Ia $ & Muscle length and speed \\ \hline
    Muscle spindle secondary & $A\beta / II $ & Muscle stretch ({\raise.17ex\hbox{$\scriptstyle\sim$}} length) \\ \hline
	Golgi tendon organ & $A\alpha / Ib $ &  Muscle contraction / tension ({\raise.17ex\hbox{$\scriptstyle\sim$}} force)\\ \hline
	Joint capsule mechanoreceptors & $A\beta / II $ &  Joint angle\\ \hline
		
	\end{tabular}
	\caption{Muscle and skeletal mechanoreceptors.}\label{tab:proprio_receptors}
\end{table*}

\subsubsection{Muscle spindles as principal proprioceptors}
Muscle spindles respond in terms of their firing rate to changes of muscle length (that is the dynamic response -- muscle spindle primary) as well to a sustained length of the muscle (static/tonic response -- muscle spindle primary and secondary). ``Position sense can therefore be envisaged as signaled by the mean rate of background discharge in muscle spindles, including that generated by both primary and secondary endings" \citep{Proske2012}. However, the position signal delivered by individual spindles seems to be far from perfect. For example: (i) muscle spindles have also efferent innervation (fusimotor -- gamma neuron activity) and thus there are differences between active and passive muscle stretching, (ii) some spindles generate background activity at all muscle lengths, no matter how short the muscle; others fall silent at short lengths. It is also not clear how the limits are retrieved and absolute position information seems to be inaccurate. Such effects may be mitigated if population of spindles are considered and possibly also by recruiting the information from joint receptors (acting possibly as limit detectors). The interested reader is referred to \citep{Jones2001,Macefield2005,Proske2012}. 

\subsubsection{From muscle length to joint angle}

\begin{figure}[htb]
	\centering
	\includegraphics[width=6cm]{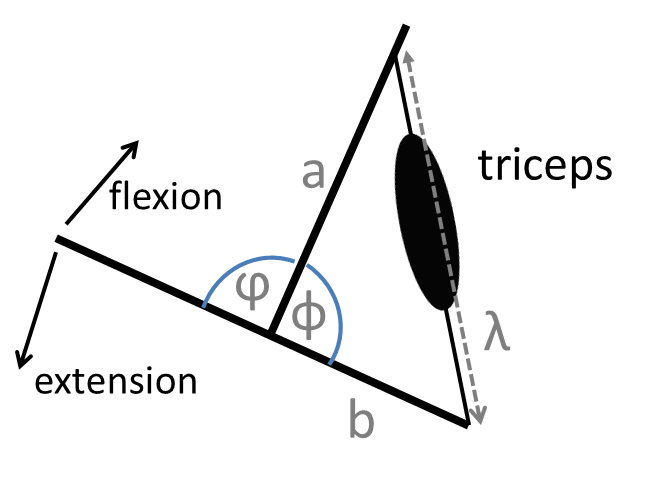}
	\caption{\textbf{Geometry of elbow joint and triceps muscle.} After \citep{Shadmehr2005}.}\label{fig:elbow_geometry}
\end{figure}

The spindles deliver information about the length of a single muscle -- flexor or extensor -- acting over a particular joint (or often also two joints in case of biarticular muscles). Not surprisingly, every spindle afferent is directionally tuned to movement of the joint in a particular direction---as can be inferred from the arrangement of its parent muscle \citep{Jones2001}. However, in order to get information about the joint angle, one has to also consider that there is a nonlinear relationship between the muscle lengths and the angle at the joint. This is illustrated in Fig.~\ref{fig:elbow_geometry} for the elbow joint (after \citep{Shadmehr2005}). The length of triceps, $\lambda$, is linked to $\theta$, the supplementary angle to the elbow angle $\phi$, by the following equation:

\begin{equation} \label{eq:joint_angle}
	\lambda = \sqrt{a^2 + b^2 - 2 \cdot a \cdot b \cdot cos(\theta)}  %a:\pi_a \rightarrow \lambda_a
\end{equation}  

Where and how (or if) the inverse of this transformation (to acquire $\theta$ and eventually $\phi$ from $\lambda$) is performed by spinal or cortical neurons has to our knowledge not been described in the literature.

To encode the 3D configuration of a complex joint (such as wrist or ankle), taking into account signals from a population of spindles of different muscles acting over the joint seems indispensable \citep{Jones2001,Bergenheim2000}---but again, it is unclear how and where would this information be combined to give an explicit representation of such 3D joint configuration.

\subsection{Neural encoding}\label{sec:neural_encoding}
As described above, there seems to be concensus that muscle spindles are the principal proprioceptors. However, it still remains unclear what the encoded variables are (in particular whether they are the lenghts of individual muscles rather than joint angles). Second, we have also not found in the literature which neural coding type is applied to the proprioceptive modality. We will focus on rate codes---where mean neural firing rate carrries information---rather than temporal (spiking) codes. The general rate types are:
\begin{itemize}
\item \textit{Intensity code.} The activity of the neuron is proportional to (or a monotonic function of) the coded variable \citep{Mallot2013}.
\item \textit{Local coding scheme.} The sensory space is divided into nonoverlapping areas that can be resolved by small groups of topographically arranged neurons with small receptive fields \citep{Ghazanfar2000}. \citet{Mallot2013} calls this \textit{channel coding} or \textit{labeled line code without overlap}. 
\item \textit{Distributed (population) coding scheme.} Neurons have relatively large and overlapping receptive fields compared with the sensory resolution measured behaviorally \citep{Ghazanfar2000}. \citep{Mallot2013} calls this \textit{labeled line code with overlap} and the ``standard way'' of neural coding.  
\end{itemize}

The encoding type is also intertwined with different \textit{neural tuning curves} that relate the average firing rate of a neuron to relevant stimulus parameters. The tuning curves cited in the literature are:
\begin{itemize}[noitemsep] 
\item \textit{Linear}
\item \textit{Sigmoidal}
\item \textit{Gaussian}
\end{itemize} 
All of the above will be implemented and illustrated later in this article in Section \ref{scc:types}. All tuning curves have specific properties which may be more suited to represent a stimulus with particular characteristics. In the linear case, firing rates are a linear function of the stimulus. Different neurons would have different equations (in terms of slope and y-intercept) (see e.g., \citep{Zipser1988}). Sigmoidal tuning curves are also monotonic but nonlinear functions of the input. Gaussian tuning curves feature a peak---are nonmonotonic functions---and may best encode the stimulus at this tuning curve peak. This may be useful if the input is noisy, since high firing rates are the neuron's most distinct response. Alternatively, stimulus may be coded not at the peak but at the high-slope regions of the tuning curve where the discrimination is the highest (see \citep{Butts2006} for an information theoretic analysis). The latter applies both to Gaussian and sigmoidal curves (not the linear functions where the slope is constant). However, we have not found in the literature what the tuning curves used for proprioceptive stimuli are.

\subsection{Ascending neural pathway}
The principal ascending neural pathway from the receptor sites to the cortex is common with that carrying ``fine touch'' and it is the so-called medial lemniscal pathway with ``relay stations'' at dorsal root ganglion, medulla, and thalamus, before reaching the neocortex, mostly terminating in area 3a.

\subsection{Representation in the cortex}
Before we briefly review the findings regarding the representation of proprioception in the cortex, it is worth noting that the findings about neuronal responses and receptive fields obtained using electrode recordings in monkeys often come from rather unspecific stimulation of so-called deep receptors: ``light to moderate taps, digit and limb manipulation, and light pressure, categorized as ‘deep stimulation', were used to stimulate the muscles, joints, and skin.``\citep{Seelke2011} A notable exception is \citep{Kim2015} who have used precisely controlled proprioceptive and cutaneous (or ``tactile``) stimulation. They also note that unlike cutaneous sensory processing, cortical mechanisms underlying proprioception are poorly understood.

\subsubsection{Anterior parietal cortex}\label{sec:APC}
The anterior parietal cortex encompasses areas 3a, 3b, 1, and 2 and is also commonly referred to as  primary somatosensory cortex (SI), even if that may be ambiguous (some consider only area 3b as primary). From these areas, area 3a is considered the most primary receptive site for proprioceptive inputs (while area 3b is the ``most primary'' cutaneous/tactile). ``Area 3a ultimately receives input from group Ia muscle spindle afferents, and contains neurons that respond to the stimulation of these and other deep receptors in the skin.`` \citep{Huffman2001}. \citep{Kim2015} precisely controlling movements of individual digits (fingers) in a macaque monkey have identified three types of proprioceptive neurons: 

\begin{itemize}[noitemsep] 
\item \textit{Single-digit position scaled}
\item \textit{Multi-digit position scaled}  
\item \textit{Posture-selective}
\end{itemize}

The first type, single-digit position scaled, means that the neuron responds in a monotonic fashion to the position (i.e. angle in a particular axis) of a single digit. This would probably be most compatible with the intensity code and a linear tuning curve. The second type, multi-digit position scaled, is sensitive to positions of mutliple digits, which is still suggestive of a rate code, but some additive mechanism for multiple joints is necessary. The posture-selective neurons, on the other hand, do not scale across specific a dimension. They are responsive to a particular posture---a specific conformation of the hand, i.e. specific position of a number of different joints. Such neurons would be compatible with either local or population coding scheme and perhaps go most naturally with Gaussian tuning curves (the peaks of tuning curves at specific positions of every joint). \citep{Kim2015} found all of these neuron types throughout all of SI: they were most abundant in area 3a, but present also in 2, 1, and 3b. While their results were obtained for the hand receptive fields only, a similar pattern may perhaps be expected also for other body parts---arm joints, for example.

Regarding topography, \citep{Krubitzer2004} using a macaque monkey found that area 3a contains a complete representation of deep receptors and musculature of the contralateral body and that the general organization of body part representations mirrors that of area 3b. However, the topographic organization is less precise and less orderly in 3a and some representations, such as the hand, are split or fractured like maps of motor cortex. 

\subsubsection{Posterior parietal areas}
Whereas in the anterior parietal cortex, the receptive fields of neurons typically span individual joints or combinations of a few adjacent joints, in the posterior parietal areas, the receptive fields are in general larger. Body posture is primarily encoded in the superior parietal lobule (areas 5 and 7), a putative homologue of the so-called parietal reach region in the monkey \citet{Pellijeff2006}.\citet{Seelke2011}, for example, analyzed the receptive field (RF) of neurons in posterior parietal area they termed 5L (lateral) of the macaque and found representation of ``deep receptors`` of the limb. The RFs of individual neurons often spanned multiple joints, such as elbow and wrist or shoulder and elbow. 

\subsection{Development in the fetus and newborn}
Somatosensation and its representation starts developing already in the womb. In fact, the  somatosensory modality starts functioning over the whole body from as early as 17th gestational age and before other sensory modalities \citet{Bradley1975}, although connections to the neocortex are not present until the 20th gestational week. Please refer to \citet{Milh2007} and references therein for a more detailed account. Spontaneous motor activity is playing an important role. From about 22 weeks, touching the mouth is already different from touching eyes (more delicate). However, \citet{Krubitzer2004} using macaque reported that unlike 3b ({\raise.17ex\hbox{$\scriptstyle\sim$}} ``primary tactile``) in which the topography seems to be present in neonates, area 3a (and 1 and 2) are not responsive in neonates in New and Old World monkeys. Furthermore, and again unlike 3b, the construction of area 3a is to a large extent based on the use of a particular body part rather than innervation density (as a matter of fact, about half of the muscle spindles in the body are located in the neck \citep{Cole1995}).

\section{Computational and robotic modeling of proprioception}
There has been some computational and robotic modeling addressing the development of somatosensory representations, even though more work has focused on the tactile rather than proprioceptive modality. The work of Kunyioshi, Mori, and colleagues has addressed prenatal development in foetus simulator
including a uterine environment. Their models can also be said to be the most biologically motivated. The foetus simulator has 198 muscles and is actuated with a central pattern generator model. It is equipped with 764 tactile sensors as well as muscle spindles and Golgi tendon organs. Different works addressed the development of SI \citep{Yamada2013} or a hierarchy of SI and SII using a denoising autoencoder \citep{Sasaki2013}. \citet{Aflalo2006} investigated motor cortex development using a Self-Organizing Map. \citet{Drix2014} studied learning proprioceptive and motor features from joint angle values in a quadruped robot using a neural ``intrinsic plasticity`` rule. Finally, \citet{Zabkar2016} presented a learning framework that builds compositional hierarchies of the joint angle space of the Nao robot.

\section{Modeling on iCub humanoid robot and Self-Organizing Maps}
We have studied the development of proprioceptive representations using the simulator of the iCub humanoid robot \citep{Metta2010,Tikhanoff2008}. Here we present an extension of the results of \citet{Bednarova2015}. 

\subsection{Data collection}\label{sec:dataset}
Data was collected during a body babbling phase. However, we did not employ random body/motor babbling, which would not induce any structure in the proprioceptive space (space of joint angles). Instead, we were inspired by \citet{Aflalo2006} who, studying the possible motor cortex development, employed so-called ethological categories: coordinated movements that the monkey uses often (e.g., hand-to-mouth, manipulation in central space in front of the monkey). Thus, we have created a simple data set where the iCub moves the hand in a region of 20x15x15 cm in front of its face and follows it with its gaze (foveating with both eyes). That is, the positions were chosen randomly in the Cartesian space and then the iCub Cartesian and gaze controllers \citep{Pattacini2011} were employed to move the hand to the position and follow it by gaze. Thus, joint positions were not randomly chosen, but corresponded to coordinated behavior inspired by a typical ``ethological category`` of an infant. Joints were sampled at 50 Hz and there were 7 arm joints (3 shoulder, 2 elbow, 2 wrist) of the left arm and 6 joints of the neck and eyes plant (neck pitch, roll, yaw; eyes tilt, version, vergence), giving 13 DoF (Degrees of Freedom) in total. The data set employed had a 20 min. duration.

\subsection{Encoding iCub joints data}\label{scc:types}
Before feeding the data to a Self-Organizing Map (SOM), we have implemented all three types of neural encoding known from biological neural systems, as reviewed in Section \ref{sec:neural_encoding}. In addition, we have also studied the case where the joint values are only normalized (taking the range of individual joints into account) and fed directly into the SOM. 

In order to determine the number of tuning curves to encode every DoF (every DoF having a different range), there are two options: (a) set the number of tuning curves constant, or (b) set the offset  between adjacent tuning curves constant. In both cases, the other variable can then be calculated correspondingly. In what follows, we have experimented with both variants. In all expressions below, $x$ represents the original angle values in degrees and $y$ represents the encoded value. The output range is in the interval $<0,1>$.

\subsubsection{Linear functions}
\begin{figure}[!h]
	\centering
	\includegraphics[width=8cm]{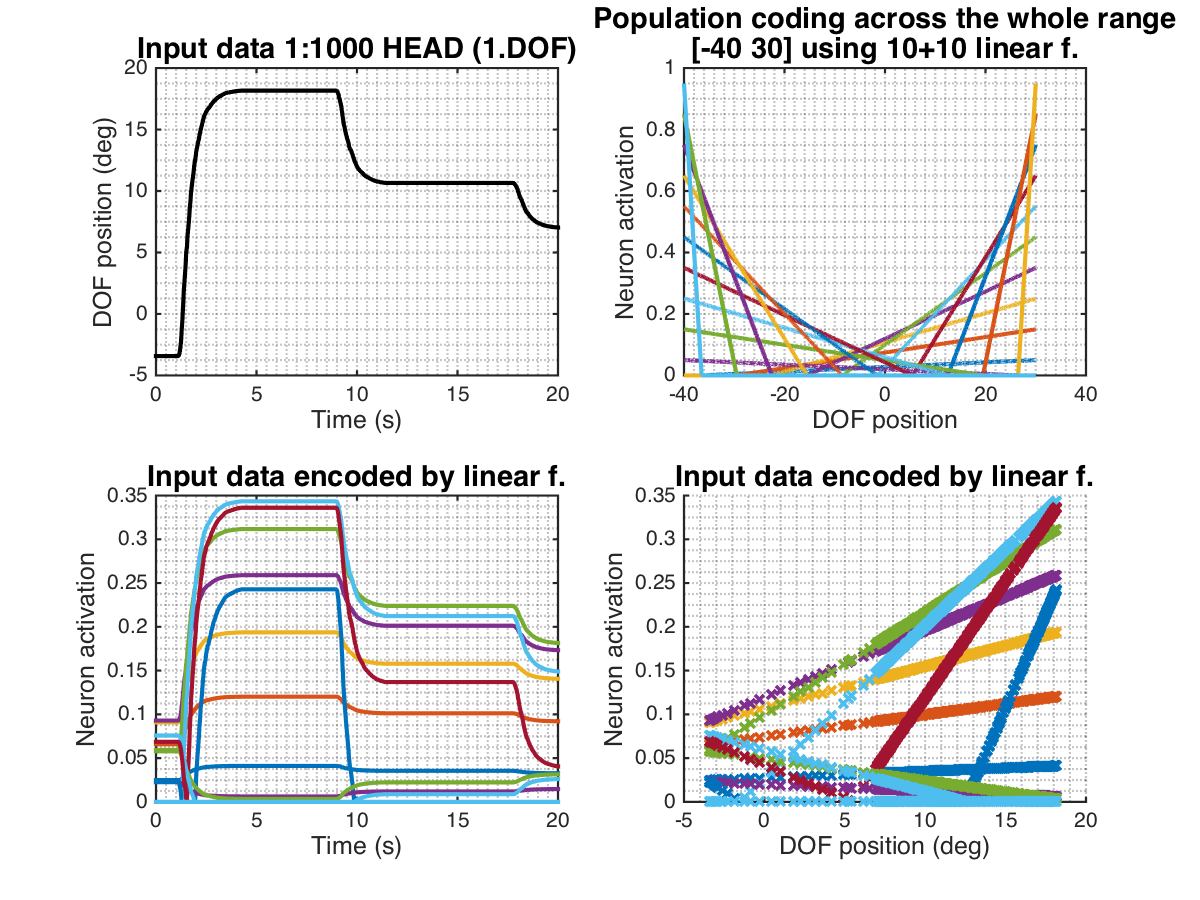}
	\caption{\textbf{Movement of one DoF encoded by linear tuning curves.} (Top left) Input data. Joint position over a 20 s interval. (Top right) Tuning curve activations on input space. (Bottom left) Input data encoded by individual tuning curves. (Bottom right) Close-up of part of the input space and its encoding.}\label{fig:13}
\end{figure}

Individual tuning curves were simply given by the equation
$$ y = (a \cdot x) + b $$
with varying parameters for the slope $a$ and y-intercept $b$. Fig.\ref{fig:13} illustrates the whole population coding for a single DoF with 20 tuning curves: 10 with positive and 10 with negative slopes.
% script z_plot13.m

\subsubsection{Sigmoids}\label{scc:sigmoids}
A sigmoidal tuning curve is given by the following equation:
$$ y =  {1 \over{1+e^{~sgn \cdot (-x+offset)}}} $$ 
where $sgn$ sets the curve orientation (positive or negative), $offset$ represents the position of the inflexion point and $x$ are the input values.

As noted above, there are two options for setting up tuning curve parameters for the population. If we set up a fixed number of curves encoding each DoF, the offsets for every DoF for sigmoids in one direction are calculated as: 
$$ offset = {(max\_range - min\_range)\over(number\_of\_sigmoids)}$$

\begin{figure}[!h]
	\centering
	\includegraphics[width=8cm]{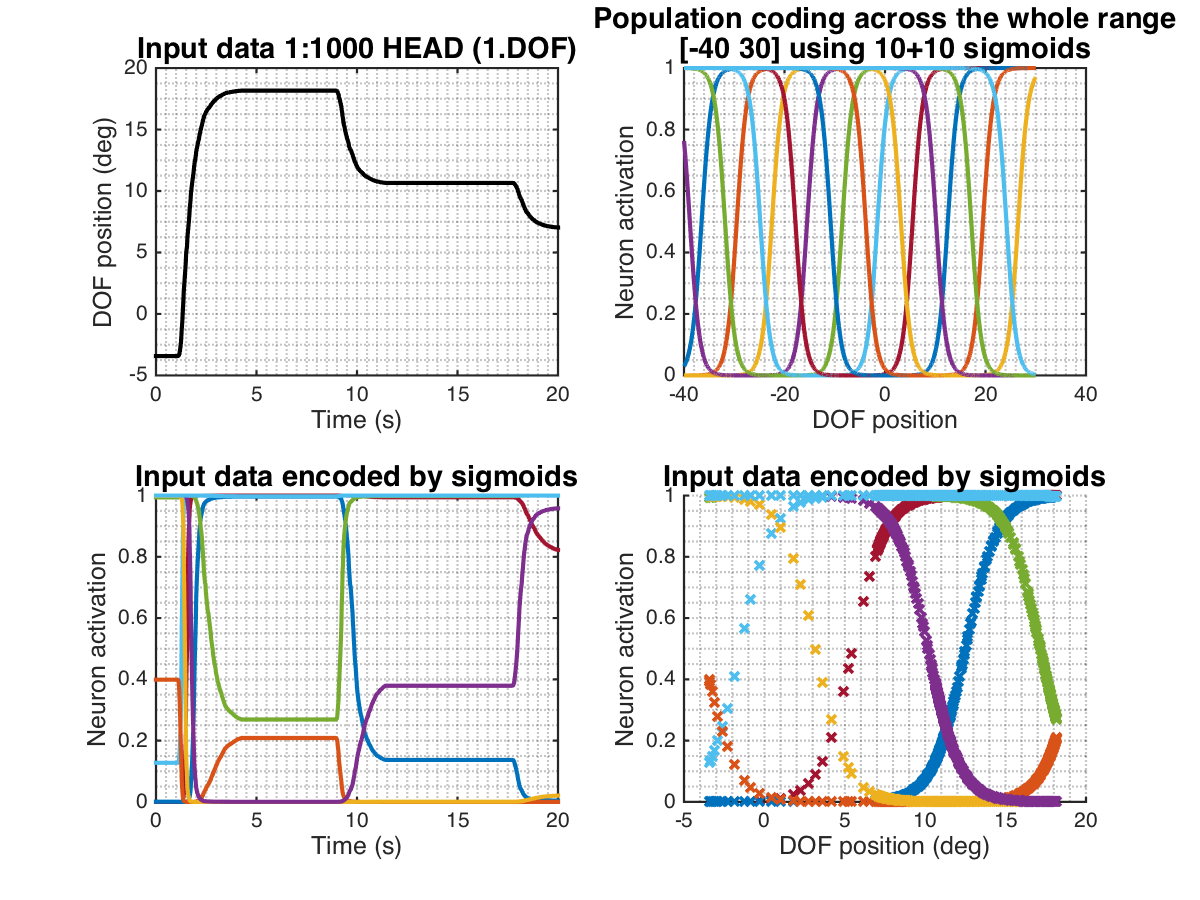}
	\caption{\textbf{Movement of one DoF encoded by sigmoidal tuning curves.} See Fig.~\ref{fig:13} for a description of all panels.}\label{fig:11}
\end{figure}

An illustration of the encoding for a single DoF is depicted in Fig.~\ref{fig:11}.

% script z_plot11.m

\subsubsection{Gaussian curves}
Gaussian tuning curves are described by the following equation:
$$ y ={e^{-{{(x-\mu)^2}\over {2\sigma ^2}}}} $$
with the curve's center or peak at $\mu$, $\sigma$ the distance to the inflexion point and $x$ the input. 

\begin{figure}[!h]
	\centering
	\includegraphics[width=8cm]{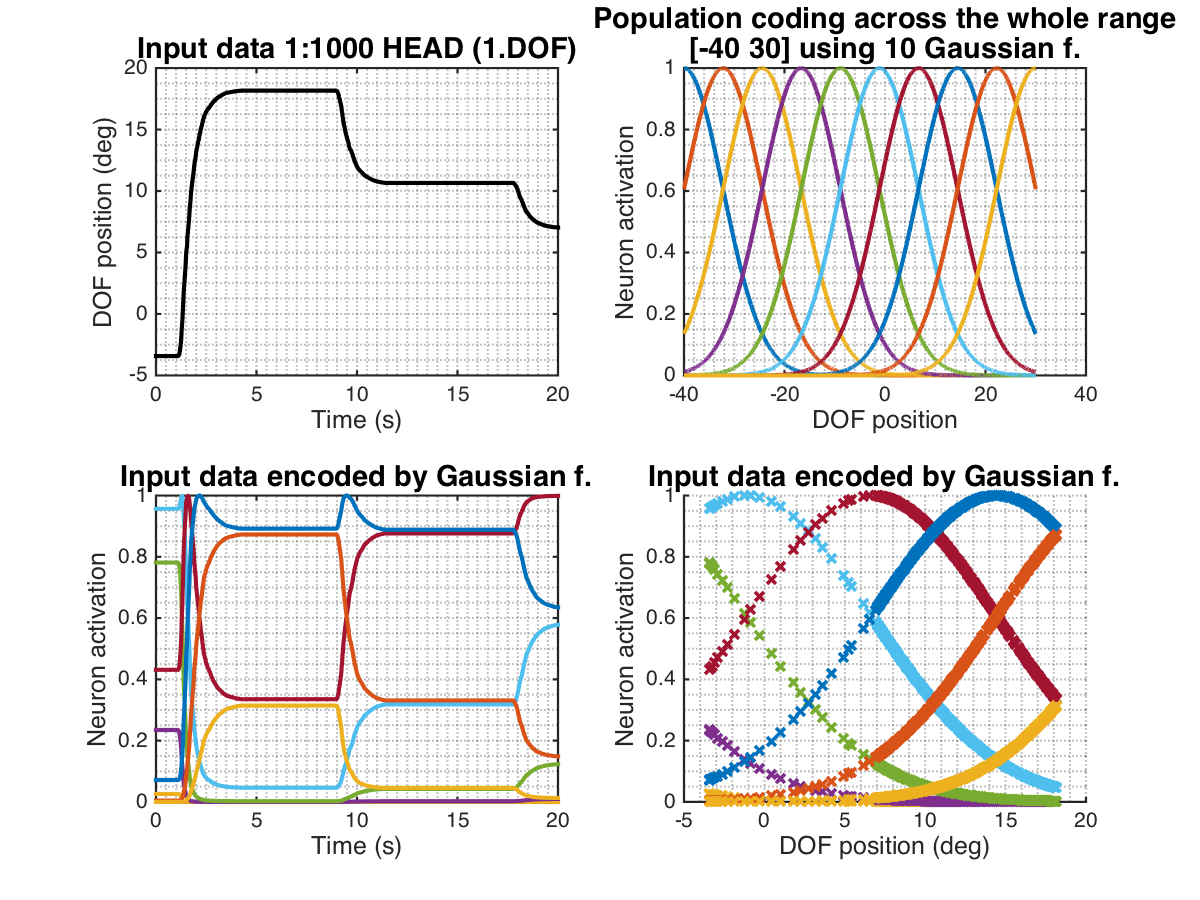}
	\caption{\textbf{Movement of one DoF encoded by Gaussian tuning curves.} See Fig.~\ref{fig:13} for a description of all panels.}\label{fig:12}
\end{figure}

In this case, the curves are symmetrical about the center and we will need only one set of tuning curves (no positive and negative). The offsets between curves can be determined similarly to the previous section. The additional parameters, $\mu$ and $\sigma$, for the population will then be calculated as follows:
$$\sigma = {(max\_range - min\_range)\over(number\_of\_gauss - 1)}$$
$$ \mu = min\_range + \sigma + offset $$
An illustration of the encoding for a single DoF is depicted in Fig.~\ref{fig:12}.

% script z_plot12.m

\subsection{Decoding}
Although the inverse mapping to the encoding may not be strictly required in all cases when stimuli are encoded by the brain, it seems that it may often be useful. In our particular application where the data are futher processed by a SOM algorithm, decoding is necessary to visualize and quantify what the network has learned. Both the linear and sigmoidal tuning curves are monotonic functions and thus a proper inverse exists. Conversely, the Gaussian tuning curve is a nonmonotonic, many-to-one function, that does not have a proper inverse. The formulas for obtaining the inverse are specified below. 
\subsubsection{Linear functions -- decoding}
$$ x = {{(y - b)}\over {a} }$$ 
\subsubsection{Sigmoids -- decoding}
$$ x = - {{ln(1-y) - ln(y) -offset \cdot sgn} \over sgn} $$
\subsubsection{Gaussian curves -- decoding}
$$ x = \mu + \sqrt{- ln(y) \cdot 2\sigma ^2} $$

\subsection{Self-organizing map on proprioceptive data}
A Self-Organizing Map (SOM) (also self-organizing feature map or Kohonen map) algorithm \citep{Kohonen1982,Kohonen1990} was used to process the encoded or normalized only joint angle data. A freely available SOM toolbox \citep{Vesanto1999_toolbox} was used.

\subsubsection{Problems with population coding as SOM input}\label{scc:problems}
Let us consider one DoF with movement range from $-40$ to $30$ degrees. Firstly, we encode the movement of this joint using 10 Gaussian curves, then we process the encoded values by a SOM, and finally we use backward transformation for decoding the result back to angles.

\begin{figure}[!h]
	\centering
	\includegraphics[width=8cm]{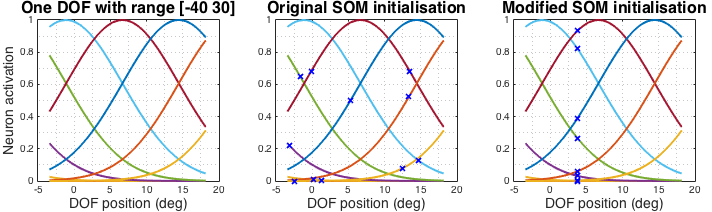}
	\caption{\textbf{Modified SOM initialisation.} See text for details.}\label{fig:21}
\end{figure}

Let us also consider a simplified version of SOM with only one output neuron. The input space after encoding has thus 10 dimensions and there are also 10 weight vectors from the input space to the single output neuron. The range of the input variables as well as weight vector components is $<0,1>$.
Standard SOM random initialisation takes all input data, calculates the range of every input dimension in the dataset and randomly picks one number from that range to initialize the weights. A property of the SOM algorithm is that the weight vector of an output neuron can be interpreted as a representative point in the input space. However, recall that the 10 dimensions are a population vector derived from a single DoF value. Hence, initializing every weight vector randomly 	will have the consequence that this point in the input space does not correspond to any angle of the joint, as illustrated in the middle panel of Fig.~\ref{fig:21} for the case of Gaussian curves. 
Therefore the initialisation of SOM had to be modified to first randomly pick an angle from the joint angle range, encode it using the respective encoding and use these values to seed the SOM weight vector---see Fig.~\ref{fig:21} right panel.

% script z_plot21.m

\begin{figure}[!h]
	\centering
	\includegraphics[width=8cm]{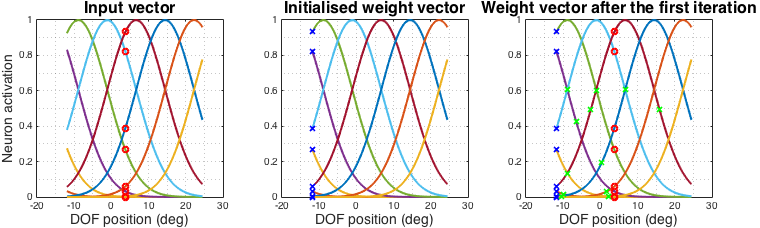}
	\caption{\textbf{Inconsistency of population coding and SOM weight vector update.}}\label{fig:22}
\end{figure}

However, this modification does not warrant that the representation will be consistent---in the sense that the representative vectors of the SOM output layer will correspond to joint angles---after learning. During training, the SOM algorithm iterates over vectors from the input space and everytime determines the most similar weight vector among the output layer neurons. Such neuron becomes the so-called best matching unit (BMU) which updates its weights in order to decrease the distance to the input presented (neurons in the BMU's neighborhood are also updated). This is illustrated in Fig.\ref{fig:22} using Gaussian curves. The left panel depicts an input vector and its encoding. The middle panel a hypothetical initialisation of the SOM weight vector to a different angle. Finally, the right panel depicts the situation after a training step -- weight vector update. The weights are drawn toward the input vector in the encoded space. However, since the different tuning curves have different slopes at the different activation values, the updated vector will not correspond to any real joint angle on the input, as illustrated by the green marks in the right panel. With Gaussian curves, there is the additional complication that the inverse mapping is not unique. This is not the case for sigmoids and linear tuning curves. However, the problem is still present, even for the linear case: even if the slope of a single curve is constant, since the pool contains lines with different slopes, the reverse transformation will also not give a unique angle of the DoF under consideration. This situation is likely to worsen with every training step. 

% script z_plot22.m

\begin{figure}[!h]
	\centering
	\includegraphics[width=8cm]{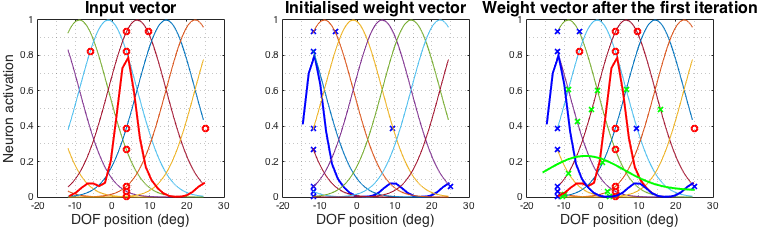}
	\caption{\textbf{Kernel density estimation for decoding the population code} from Fig.~\ref{fig:22}.}\label{fig:23}
\end{figure}

In this case, if one wants to decode from the SOM's weight vector to the original space of joint angles, that is decode the values of a population, there is no straightforward solution. The method that worked the best for us was non-parametric Kernel Density Estimation \footnote{\url{http://www.mathworks.com/help/stats/kernel-distribution.html}} (KDE) that estimates the probability distribution of input data using the following expression: 
$$f_h(x)= {1\over{nh}} \sum_{i=1}^{N}K({{x-x_i}\over{h}});~-\infty<x<\infty $$
where $h$ represents so called 'bandwidth' -- width of the window, influencing the 'smoothness' of the curve, $K_h$ represents normalised probability function, for example Gaussian, and $x_i, i=1..n$ is the input data. An example is provided in Fig.~\ref{fig:23}.

% script z_plot23.m

\subsubsection{Training a 5x5 SOM from the 10 DoF data set}
Using the data set specified in Section \ref{sec:dataset}, we trained a SOM with a 5x5 output lattice.  Various combinations of encoding (linear, sigmoid, Gaussian curves, or normalized inputs only), order of inputs as presented during training (ordered as in data set or randomly shuffled), and training length (number of reiterations of the whole input dataset in one map training) were tested. The connections from the input layer to the SOM were all-to-all, that is, every output neuron of the map was learning to represent a point in the space of all DoF---a complete posture of the robot in terms of arm and head. 

'Random' input data order as well as longer training length always gave us better results, so we will report results for this setting only. %compared to 'ordered' input and shorter training. 
We also tested the two variants of population coding setup---fixed number of curves per DoF or fixed offset between curves. Tests showed that the resulting map has better quality if we set up constant number of functions encoding every DoF. This guarantees that every DoF will have an equal number of input neurons to the SOM.

\begin{figure}[!h]
	\centering
	\includegraphics[width=8cm, height=7cm]{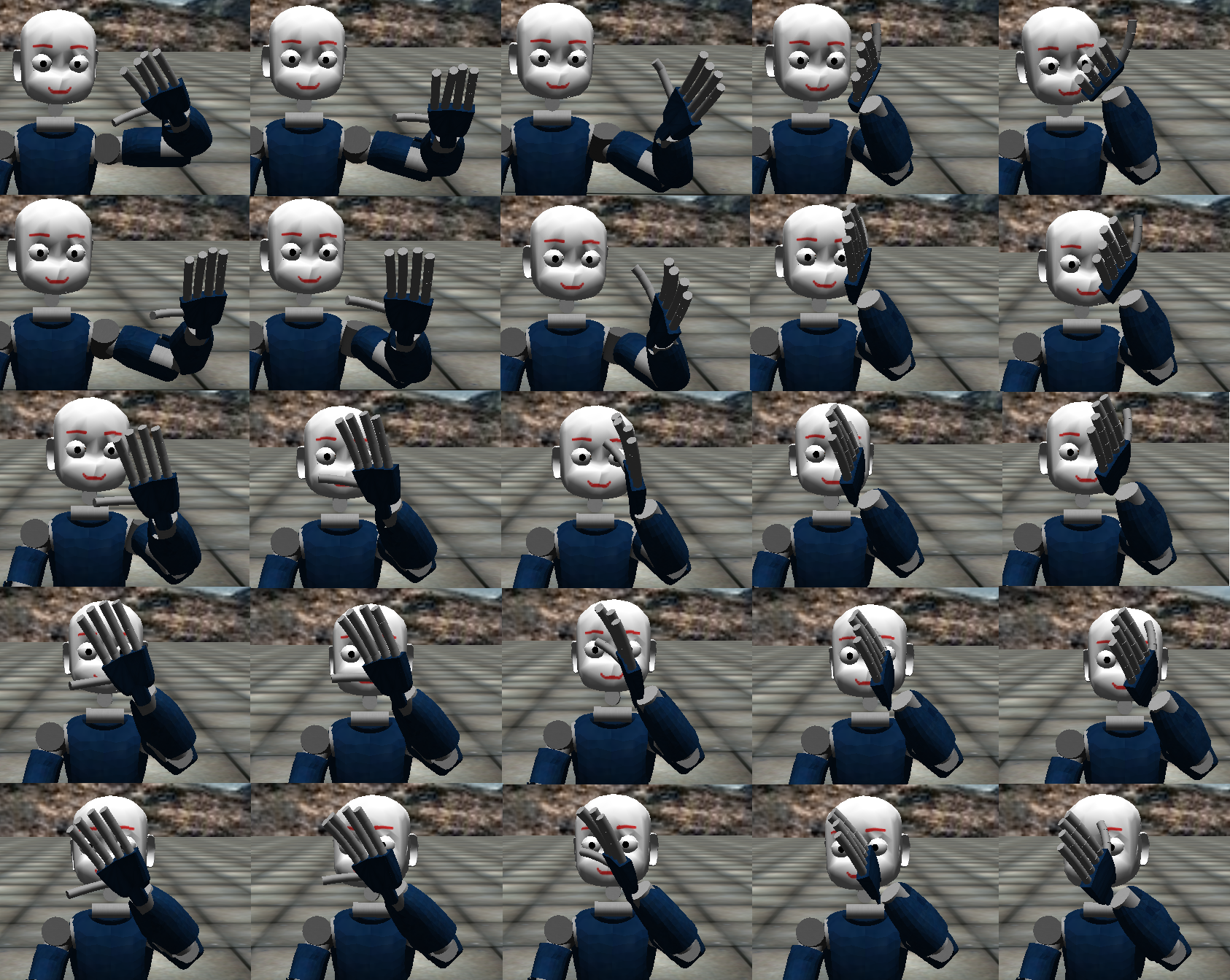}
	\caption{\textbf{Visualization of SOM after training and decoding to angles.} Every cell of the 5x5 matrix illustrates the posture represented by each of the output layer neurons. The decoded DoF positions were visualized on the iCub simulator.}\label{fig:41}
\end{figure}

Various measures have been proposed to numerically assess the organization of trained SOMs (for an overview, see \citep{Polani2002} and references therein). In general, they either assess the vector quantization or the topology preservation capability of the SOM algorithm. We have experimented with different measures. First, Fig.~\ref{fig:41} provides a qualitative insight by visualizing one of the best learned maps (with Gaussian tuning curves). The representative vectors of every output neuron were decoded back to the joint angle space and visualized on the iCub simulator. Some topology preservation in the posture space is apparent.

\begin{figure}[!h]
	\centering
	\includegraphics[width=7cm]{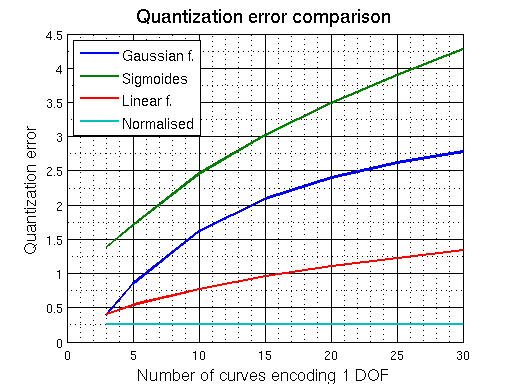}
	\caption{\textbf{Comparison of quality of learned maps in terms of quantization errror.} Results for different encoding types as well as different number of tuning curves are plotted. For the sigmoids and linear functions, the nr. curves per DOF corresponds to their count with one orientation---the total number was then double.  Random input order and training length of 6 cycles was used in all cases.}\label{fig:5}
\end{figure}

A quantitative insight is provided in Fig.~\ref{fig:5}. Quantization error measures the quality of input data representation by calculating the average distance of every input vector with its BMU vector from the learned map. We have evaluated the criterion for different types of population coding and different number of tuning curves. In addition, we tested the case in which the joint angle values are only normalized and passed directly to the SOM. Note that in this case, the input space has only 10 dimensions (10 DoF), whereas in the other cases where population coding is applied, the input space dimension is 10 x nr. curves encoding a single DoF. 

The results clearly demonstrate that all population encoding variants lead to worse results compared to the case where normalized inputs were directly fed to the SOM. This may be surprising at first, but it is a consequence of the problems illustrated in Section \ref{scc:problems}. The problem is amplified if more tuning curves per DoF are used. Furthermore, the situation gets also worse for the nonlinear tuning curves (sigmoids and Gaussians). This confirms that there is an inherent incompatibility between population coding and the adaptation step of the SOM algorithm.

% figure; subplot(1,2,1); z_plot51; hold on; subplot(1,2,2); z_plot52;

%%%%%%%%%%%%%%%%%%%%%%%%%%%%%%%%%%%%%%%%%%%%%%%%%%%%%%%%%%%%%%%%%%%%%%%% CHAPTER 4
\section{Discussion, open questions, future work}
This article without doubt raises more questions than provides answers. With the goal of providing a computational model of the representation of proprioception in the brain and its development, it falls significantly short of its aspirations. However, we feel that a number of important issues have been identified, which we discuss below. 

First of all, empirical knowledge regarding the workings and development of proprioception and its representation in biology is fragmented and partially contradictory. Or, perhaps, such is the information that the the neural systems dealing with proprioception are processing. \citet{Feldman2009}, for example, concludes that ``ambiguity of positional information is characteristic of all afferent signals involved in kinesthesia``. We want to make a number of observations. First, the view of isolated bottom-up processing of proprioceptive information is wrong. There is now sufficient evidence suggesting that processing somatosensory information is a ``team effort" \citep{Saal2014}, where tactile and proprioceptive \citep{Kim2015} and even vestibular and downstream (top-down) information \citep{Huffman2001} is integrated. Area 3a---even if the most ``proprioceptive`` and most primary---can thus hardly be considered as an isolated area processing ascending proprioceptive inputs. Second, the mechanism of formation or development of the proprioceptive representations is unclear. As pointed out earlier, there is evidence suggesting that the topography of area 3a in monkeys is to a large extent learned after birth \citep{Krubitzer2004}. However, once learned, its topography is very similar to that of neighboring area 3b receiving primary tactile inputs. This may seem surprising, since the ``content'' of information coming from the two modalities seems quite different. Nonetheless, the connections between these areas and probably also the frequent correlations in proprioceptive and tactile inputs from same body parts could account for this. 

The computational model we presented rests on a number of simplifying assumptions. First, the data set used is preliminary and arbitrary: we used only one coordinated movement type---or ethological category \citep{Aflalo2006}---where the robot was moving the hand in front of its face and following it with gaze. The repertoire of behaviors in newborns needs to be investigated and implemented. Furthermore, newborns' coordination may in fact be severely limited and it is also possible that the appropriate stimulation type is not coordinated movement, but rather more random and isolated movements such as the muscle twitches \citep{Khazipov2004}. The correlation structure in the proprioceptive data could be induced through the agent's embodiment. This would probably require a more detailed model of the musculo-skeletal system as well as implementation of reflexes such as the stretch reflex (similar to \citep{Kuniyoshi2006}). Second, we did not deal with all the submodalities of proprioception and restricted ourselves to ``muscle spindles``---in fact, taking an additional shortcut and directly using joint angles rather than the length of (possibly also biarticular) muscles. The typical antagonistic arrangement of muscles (flexor, extensor pairs) was somewhat emulated when tuning curvers in opposite directions were considered. 

In our model, we experimented with different types of population coding and the Self-Organizing Map algorithm. The most important findings are detailed in what follows. First, as detailed above in Section~\ref{sec:APC}, \citet{Kim2015} identified different types of proprioceptive neurons in SI, namely position-scaled (single or multi-digit) and posture-selective. It is evident that our model can only learn to represent the posture-selective type. The SOM algorithm is in essence learning to efficiently represent points from the input space, devoting more space in the output layer to more frequent inputs and carrying over topological relationships from the input to the output space. In our case, it basically learns to pick up the most frequent postures (or ``postural synergies``). However, it cannot preserve the position-scaling property---a type of intensity code---increasing neuron firing  as a monotonic function of joint position. Finding an algorithm that would preserve this feature while at the same time having the useful properties of SOM (vector quantization, topology preservation) remains the topic of our future work. Second, in our model, the connections from the input space (joint angles, possibly after population coding) to the output layer were all-to-all. That is, every output neuron of the SOM was learning to represent a configuration of all arm and head joints. This is problematic from a mathematical point of view because the topological relationships in the input space---similarity of the different postures in the configurations of all 10 DoF---are clearly of higher dimensionality than 2, while the output layer lattice of the SOM we used was only 2-dimensional and thus the algorithm was faced with the impossible task of reducing the topological manifold from the input space to mere 2 dimensions. Instead, the RFs of neurons in anterior parietal cortex typically span single or multiple joints \citep{Kim2015}, or multiple joints (e.g., 2 or 3) in the posterior parietal cortex \citep{Seelke2011}. Therefore, it seems that the connections between input and output should not be all-to-all but constrained to some regions of the input space only. A modification of the SOM algorithm constraining the maximum receptive field size (MRF-SOM) of the output neurons was presented in \citep{Straka2014,Hoffmann2016}. A hierarchy of such maps may give rise to similar gradual expansion of RFs as observed when going upstream in the cortex. Third, we have learned from our study that the SOM algorithm has intrinsic difficulties with input preprocessing using population code. This has to do with the fact that the input (joint angle in this case) is encoded using a population of tuning curves that have different slopes (different in different lines in the linear function case; different even at different points of single curve in sigmoids and Gaussians). After encoding, the SOM weight update occurs in this transformed space; however, this leads to inconsistency in the population if it is to be mapped back to the initial joint space. This is even more problematic if Gaussian curves are used as there is no proper inverse due to their nonmonotonicity. Perhaps, it may not be the necessity for the brain to peform this decoding; yet it seems that the encoding in combination with the SOM adaptation step does introduce undesired transformations to the information that is represented. In the robot, where joint angles are directly available, using them directly leads to superior results.  In biological systems, population coding may be inevitable because the information is only available in a distributed fashion, such as from populations of muscle spindles and other receptors. As a consequence, neural learning mechanisms may need to work in different ways than the classical SOM in this case.

%%%%%%%%%%%%%%%%%%%%%%%%%%%%%%%%%%%%%%%%%%%%%%%%%%%%%%%%%%%%%%%%%%%%%%%% ACKNOWLEDGEMENT
\vspace{5ex}
\noindent
{\bf\large Acknowledgment}\vspace{2ex} \newline
{
MH was supported the Marie Curie Intra European Fellowship (iCub Body Schema 625727) within the 7th European Community Framework Programme. We are also indebted to Igor Farkaš for valuable comments and discussions. } 

%%%%%%%%%%%%%%%%%%%%%%%%%%%%%%%%%%%%%%%%%%%%%%%%%%%%%%%%%%%%%%%%%%%%%%%% BIBLIOGRAPHY
\vspace{3,16314mm}
\nocite{*}
% apa - style
\bibliographystyle{apalike_kuz-en}
\bibliography{HoffmannBednarova_KUZ2016}

%% citacie ulozte do suboru references.bib
%% na populaciu zoznamu literatury pouzite program
%%
%% bibtex references
%%
%% po ktorom je potrebne dokument znova zlatexovat

\end{document}